\documentclass[11pt]{article}
\usepackage{float}
\usepackage[final]{acl}
\usepackage{amssymb}
\usepackage{times}
\usepackage{latexsym}
\usepackage{amsmath}
\usepackage[T1]{fontenc}
\usepackage{multirow}
\usepackage[utf8]{inputenc}
\usepackage{booktabs}
\usepackage{microtype}
\usepackage{listings}
\usepackage{xcolor}
\usepackage[ruled,vlined]{algorithm2e}
\lstdefinestyle{promptstyle}{
  basicstyle=\ttfamily\footnotesize,
  backgroundcolor=\color{gray!10},
  frame=single,
  framesep=4pt,
  breaklines=true,
  breakatwhitespace=false,
  showspaces=false,
  showstringspaces=false,
  columns=fullflexible,
  keepspaces=true,
  captionpos=b,
}
\usepackage{inconsolata}

\usepackage{graphicx}
\usepackage{makecell}
\title{\includegraphics[width=0.05\textwidth]{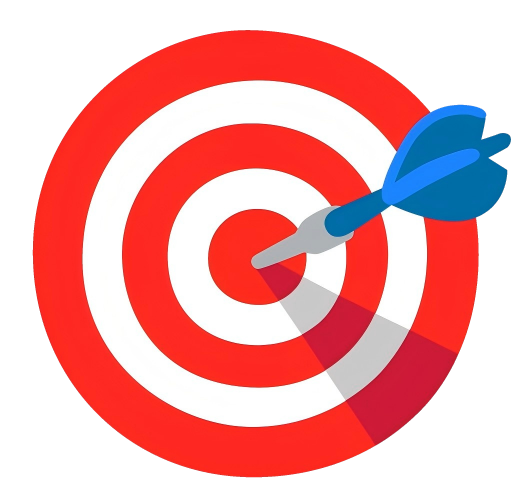}VAA-CSEC: Vote-guided Advantage Allocation for Chinese Semantic Error Correction}

\author{
  \textbf{Yitong Han\textsuperscript{1}},
  \textbf{Nankai Lin\textsuperscript{1,2}}$^{\dagger}$,
  \textbf{Juan Luo\textsuperscript{1}},
  \textbf{Hongyan Wu\textsuperscript{3}},
  \textbf{Lianxi Wang\textsuperscript{1,2}},
  \textbf{Shengyi Jiang\textsuperscript{1}}
\\
  \textsuperscript{1}School of Information Science and Technology, Guangdong University of Foreign Studies
  \\
  \textsuperscript{2}Guangdong Engineering Research Center of Data Security Governance and Privacy Computing
  \\
  \textsuperscript{3}College of Computer Science and Technology, National University of Defense Technology
\\
  \small{
    $^{\dagger}$Corresponding author}
\\
  \small{20231003317@mail.gdufs.edu.cn, neakail@outlook.com}
      }

\begin{document}
\maketitle
\begin{abstract}
Chinese Semantic Error Correction (CSEC) targets semantic errors in Chinese text, which are typically more subtle and complex than spelling and grammatical errors but remain relatively underexplored. Existing LLM‑based approaches face two recurring obstacles in this task: over‑correction, and unclear interaction between Chain‑of‑Thought (CoT) reasoning and self-consistency decoding, such that the benefits brought by CoT cannot be reliably transferred to final corrections. We propose \textbf{V}ote-guided \textbf{A}dvantage \textbf{A}llocation for \textbf{CSEC} (\textbf{VAA-CSEC}), a multi-stage framework that combines CoT distillation, Supervised Fine-Tuning (SFT), Reinforcement Learning (RL) and self-consistency decoding. During RL, we design a task-specific reward function that directly aligned with the minimal-editing principle of CSEC. We further introduce \textbf{G}roup-\textbf{L}evel Relative \textbf{P}olicy \textbf{O}ptimization (\textbf{GLPO}), which reallocates GRPO advantages according to the margin between individual rollout rewards and the vote-aggregated group reward, aligning the RL training objective with the self-consistency objective used at inference time. Experiments on CSED-C and NaSGEC-Exam show that VAA-CSEC outperforms all LLM-based baselines on CSED-C with an $F_{0.5}$ of 47.72\%, achieves the highest recall of 42.15\% among all methods, and establishes a new state of the art of 41.55\% $F_{0.5}$ on NaSGEC-Exam.\footnote{Code is released at https://github.com/HanYiton/VAA-CSEC}
\end{abstract}

\section{Introduction}\label{introduction}
Text Error Correction (TEC) is a fundamental task in Natural Language Processing (NLP), which aims to automatically detect and correct errors in text. According to the type of error involved, TEC is typically divided into three categories, spelling error correction \cite{BIJOY2025101703,jiaprobability}, grammatical error correction \cite{marier2025grammatical,FANG2025127397} and semantic error correction \cite{WU2026131925}\footnote{Detailed examples of the three tasks can be found in Appendix \ref{example}.}. For Chinese, the spelling and grammatical correction have achieved substantial progress, but Chinese Semantic Error Correction (CSEC) remains underexplored due to the subtlety of semantic errors and the difficulty of correction, despite their potential to introduce ambiguity and misinterpretation in downstream applications.

\begin{figure}
    \centering
    \includegraphics[width=1\linewidth]{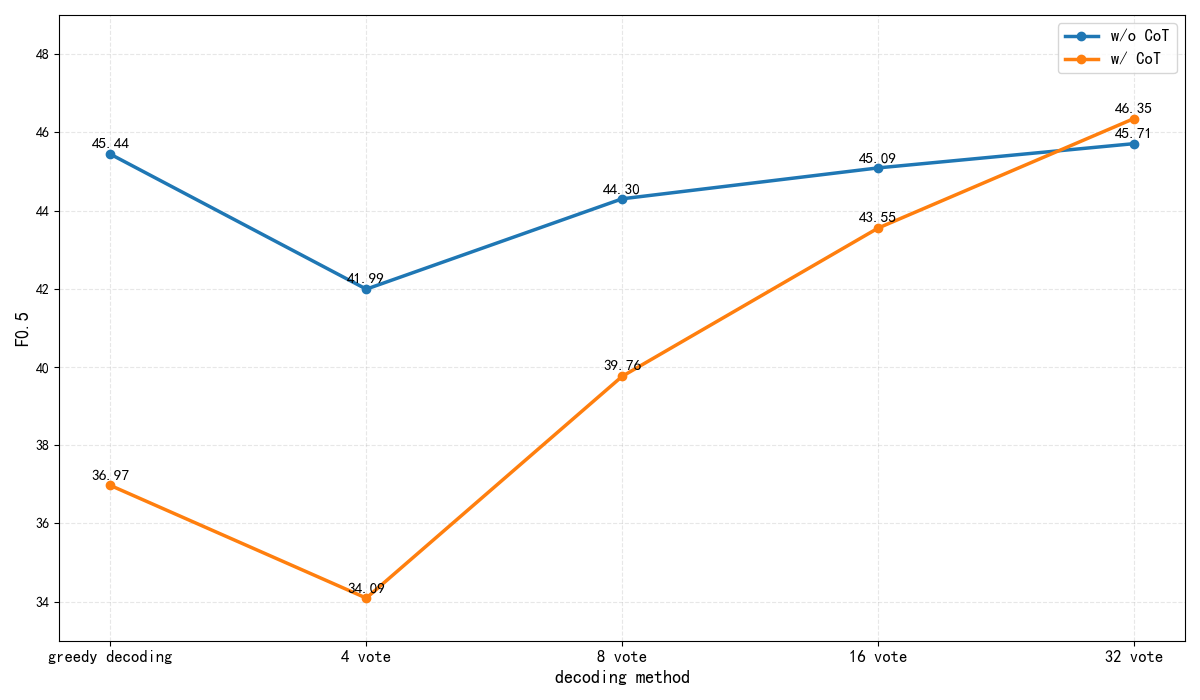}
    \caption{Comparison of $w$/ CoT and $w/o$ CoT models. We train the same model twice separately: once on CoT data and once on non-CoT data.}
    \label{fig:compare}
\end{figure}

Recent work on Chinese Text Error Correction (CTEC) has increasingly adopted Large Language Models (LLMs), owing to their strong ability to generate fluent and coherent text \cite{WU2026131925,liu-etal-2025-chain,chang-zhu-2025-chunk}. However, two fundamental challenges remain unresolved. First, text correction inherently follows the principle of minimal editing, whereas LLMs often introduce unnecessary modifications to originally correct spans or even rewrite entire sentences, resulting in severe over-correction. Second, although Chain-of-Thought (CoT) reasoning and self-consistency decoding \cite{wang2022self} have shown promising gains in reasoning tasks, recent studies suggest that individual reasoning paths generated by LLMs may be unreliable and not faithfully grounded in the input context \cite{ye2022unreliability}. Consequently, the role of CoT under different decoding strategies in correction tasks remains poorly understood. Our preliminary experiments further proves this phenomenon (Figure~\ref{fig:compare}): under greedy decoding, $w$/ CoT model perform worse than $w/o$ CoT model, yet under self-consistency, the $w/o$ CoT model model gains plateau while $w$/ CoT model improve steadily and eventually surpass it. This suggests that the benefit of CoT in CSEC is fundamentally tied to the sampling strategy, and that training objectives explicitly aligned with self-consistency are needed to fully exploit CoT-based models.

To address these two challenges, we propose \textbf{V}ote-guided \textbf{A}dvantage \textbf{A}llocation for \textbf{CSEC} 
(\textbf{VAA-CSEC}), a multi-stage framework that combines CoT distillation, Supervised Fine-Tuning (SFT), 
Reinforcement Learning (RL), and self-consistency decoding. At the RL stage, we design a task-specific reward function with two 
components: a format constraint that regularizes the output structure, and an edit-distance-based correctness term that rewards effective corrections while explicitly penalizing over-correction, directly operationalizing the minimal-editing principle of CSEC. We further propose \textbf{G}roup-\textbf{L}evel Relative \textbf{P}olicy \textbf{O}ptimization (\textbf{GLPO}), which reallocates GRPO advantages according to the margin between individual rollout rewards and the vote-aggregated group reward, aligning the RL training objective with the self-consistency objective used at inference time. Our contributions are as follows.
\begin{itemize}
\setlength\itemsep{1pt}
\item We propose \textbf{VAA-CSEC}, a multi-stage framework that couples CoT-based reasoning with self-consistency-aware RL to address the unique challenges of identifying and minimally correcting subtle semantic errors in Chinese text.
\item We design a task-specific reward function for CSEC that jointly enforces output format, rewards effective edits, and penalizes over-correction, directly operationalizing the minimal-editing principle.
\item We introduce \textbf{GLPO}, an RL strategy based on GRPO that aligns training-time optimization with inference-time self-consistency by reshaping advantages according to a group-level voting reward.
\item Experiments show that VAA-CSEC achieves the best performance among all LLM-based methods on CSED-C with an $F_{0.5}$ of 47.72\% and the highest recall of 42.15\%, and establishes a new state of the art on NaSGEC-Exam with an $F_{0.5}$ of 41.55\%.
\end{itemize}

\section{Related Work}

\subsection{Traditional Chinese Text Error Correction Methods}
CTEC has evolved from rule-based and statistical methods to deep learning approaches. Early work relied on manual rules and statistical models such as N-gram language models and confusion sets~\cite{Lin-and-Chu-2015-gector}, but these methods struggled with complex, context-dependent errors~\cite{Dahlmeier-and-Ng-2011-gector}. Neural approaches then became mainstream, forming two paradigms: 
Seq2Seq, which casts correction as translation from incorrect to correct sentences~\cite{Zhao-and-Wang-2020-gector,Rothe-etal-2020-gector}, and Seq2Edit, which predicts position-wise edit tags~\cite{Omelianchuk-etal-2020-gector}.

More recently, LLMs have been widely adopted for CTEC and achieved strong results, but they still suffer from over-correction. \citet{li2025harnessingrulebasedreinforcementlearning} report that on Chinese grammar correction and spelling check tasks, LLMs still trail traditional deep learning methods on several metrics, making the mitigation of over-correction an important open problem. In parallel, semantic correction remains underexplored relative to spelling and grammatical correction due to the hidden nature of its errors and the difficulty of corpus annotation~\cite{Sun-etal-2023-gector}, 
which motivates dedicated study of this setting.

\subsection{Text Error Correction Methods Using Reinforcement Learning}
RL is increasingly used in the post-training stage of NLP tasks. Within the Reinforcement Learning from Human Feedback (RLHF)~\cite{Kaufmann-etal-2023-gector}, the 
dominant algorithms have evolved from PPO~\cite{Schulman-etal-2017-gector} and DPO~\cite{Rafailov-etal-2023-gector} to GRPO~\cite{shao2024deepseekmathpushinglimitsmathematical}.

Applying RL to LLM-based text correction has also progressed rapidly. \citet{li2025harnessingrulebasedreinforcementlearning} combine GRPO with rule-based rewards for Chinese grammar correction, achieving state-of-the-art performance on FCGEC with markedly improved recall. To address advantage collapse under sparse rewards, EDGE-GRPO~\cite{Zhang-etal-2025-gector} introduces entropy-driven advantage estimation and a guided correction mechanism. \citet{Li-etal-2026-gector} further stabilize training via 
bilateral contextualization and reward confidence correction, and \citet{Lin-etal-2026-gector} propose CEC-Zero, a zero-supervised RL framework that uses semantic similarity and candidate clustering to compute consensus rewards. Together, these results show that combining RL with LLMs is an effective path toward mitigating over-correction, improving recall, and strengthening cross-domain generalization in text correction.

\section{Method}

\subsection{Overview}
\begin{figure*}[htbp]
    \centering
    \makebox[\linewidth]{%
        \includegraphics[width=\linewidth]{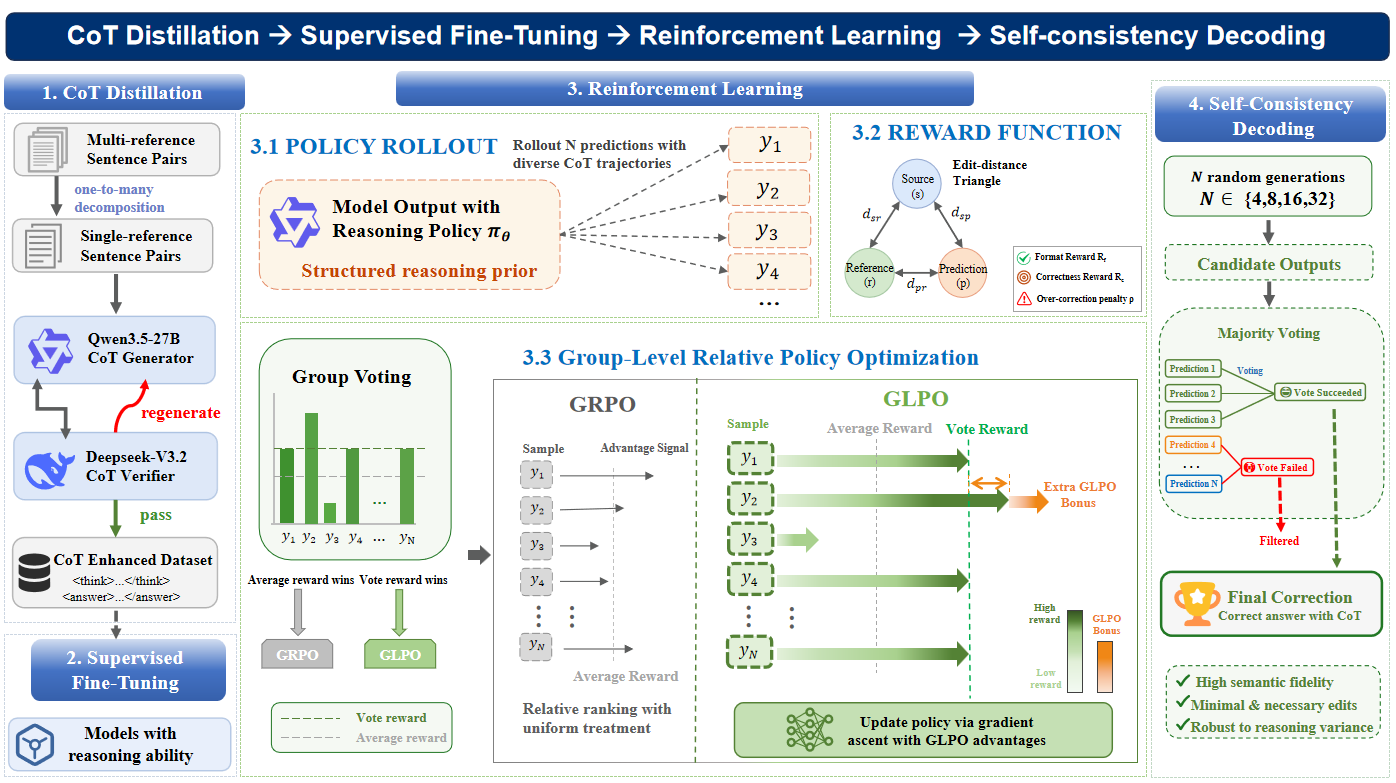}%
    }
    \caption{Overview of VAA-CSEC.
\textit{Left:} a CoT distillation pipeline pairs a Qwen3.5-27B generator with a DeepSeek-V3.2 verifier to produce structured data for SFT.
\textit{Center:} GLPO rolls out $N$ candidates from the policy $\pi_\theta$, scores them with a reward combining format $R_f$, edit-distance correctness $R_c$, and an over-correction penalty $\rho$, and reallocates advantages using the margin between each rollout's reward and the vote-aggregated group reward. \textit{Right:} at inference, $N$ stochastic generations are aggregated by majority voting to yield the final correction.}
    \label{fig:framework}
\end{figure*}

Our VAA-CSEC consists of four stages: CoT distillation, SFT, RL, and self-consistency decoding. Figure~\ref{fig:framework} shows the detail of our VAA-CSEC framework. We unfold training data with multiple references and use Qwen3.5-27B\footnote{https://huggingface.co/Qwen/Qwen3.5-27B} to distill CoT rationales for each sentence pair, with DeepSeek-V3.2\footnote{https://huggingface.co/deepseek-ai/DeepSeek-V3.2} acting as a quality reviewer (§\ref{sec:distillation}). In the SFT stage, we fine-tune Qwen3.5-4B\footnote{https://huggingface.co/Qwen/Qwen3.5-4B} on the CoT data to obtain a reasoning model (§\ref{sec:sft}). Then we design a task-specific reward for CSEC and further propose GLPO, a strategy that aligns RL training with self-consistency at inference by reshaping advantages according to a group-level voting reward (§\ref{RL}). During inference, we perform self-consistency decoding over N stochastic generations (§\ref{nvote}). 

\subsection{Task Definition}
Given a Chinese text sequence $X=[x_1, x_2, \dots, x_n]$ that may contain semantic errors, the CSEC task aims to develop a system that transforms it into a semantically correct and naturally expressed text sequence $Y$, while preserving the core semantics of the original sentence and making only minimal changes to the necessary parts. Since this process is essentially a mapping from erroneous text to corrected text, it can be viewed as a special form of text translation.

\subsection{CoT Distillation}\label{sec:distillation}
To equip the model with basic reasoning and correction capabilities before RL, we construct CoT rationales for each sentence pair in the training set through an automatic distillation-verification pipeline. We first unfold training data with multiple references, so that each source sentence corresponds to only one reference sentence. For every such pair, we prompt a strong teacher model, Qwen3.5-27B, to generate a CoT rationale that explains how the source can be revised into the reference. To filter low-quality rationales, we select DeepSeek-V3.2 to verify each generated CoT. Samples that fail verification are regenerated, up to three attempts. The processed data is shown in Experiments \ref{dataset}, and the full generation and verification prompts are provided in Appendix~\ref{A_template}.

\subsection{Supervised Fine-Tuning}\label{sec:sft}
Based on the distilled CoT data, we perform SFT on Qwen3.5-4B with LoRA, obtaining a SFT checkpoint that serves as the initialization for the subsequent RL stage. The SFT stage plays two essential roles in our pipeline. First, it teaches the model to follow the structured <think>\dots</think> <answer>\dots</answer> output format, which is a prerequisite for our reward function to reliably extract the final correction during RL. Second, it grounds the policy in the distribution of high-quality reasoning trajectories distilled from the teacher, so that RL exploration starts from a competent rather than a random policy.

\subsection{Reinforcement Learning}\label{RL}
The observation in Introduction \ref{introduction} is consistent with recent studies~\cite{ye2022unreliability}, which show that individual reasoning chains generated by LLMs may themselves be unreliable, while correctness often emerges from agreement across multiple sampled reasoning paths. For a CoT-based correction model whose $N$ sampled candidates already contain the correct answer with high probability, the primary bottleneck of self-consistency is therefore no longer the model's intrinsic correction capability, but how reliably the correct correction is reproduced across sampled candidates and ultimately dominates the vote. Motivated by this perspective, we introduce GLPO on top of the SFT 
stage to align the RL training objective with the self-consistency objective at inference time, encouraging correct corrections to be reinforced as the dominant mode across sampled candidates.

\subsubsection{Reward Function}
For CSEC, the reward function should not only judge whether the model successfully corrects the errors in the sentence, but also further evaluate the quality and rationality of the revisions made by the model. 
Several studies have explored this issue. For example, \citet{li2025harnessingrulebasedreinforcementlearning} proposed a rule-based reward for CGEC. However, it fails to distinguish two sub-cases of "originally incorrect, modified but still incorrect": whether the output moves closer to or further from the reference. These cases carry very different optimization implications—partial correction versus semantic damage and assigning them the same reward deprives the model of fine-grained learning signals.

Therefore, we propose a reward function design method inspired by the $F_{0.5}$ scoring scheme. Our reward function consists of two components: format reward and correctness reward.

\noindent\textbf{Format Reward.} The format reward is designed to constrain the model to generate outputs in the standard structured format: <think>...</think> <answer>...</answer>. Additional rewards are assigned only when the model strictly follows the specified format. This design can stabilize the output format of the model during the RL stage and enhance the controllability of both the generated reasoning process and the final answer.

\noindent\textbf{Correctness Reward.}
For the correctness reward, we compute three edit distances:
$d_{sr}$ between the source and the reference (the essential edits required),
$d_{sp}$ between the source and the model output (the actual edits performed),
and $d_{pr}$ between the model output and the reference (the residual gap).
We define the number of useful edits as
\begin{equation}
u = \frac{d_{sr} + d_{sp} - d_{pr}}{2},
\label{eq:useful}
\end{equation}
which captures the portion of the model's edits that genuinely move the output toward the reference. When $u > 0$, the edits are beneficial, when $u = 0$, they yield no net gain, when $u < 0$, the output drifts further from the reference than the source itself, in which case a negative reward is assigned. Based on $u$, we define edit-based precision and recall as
\begin{equation}
P = \min\!\left(\frac{u}{d_{sp}},\,1\right),
\end{equation}
\begin{equation}
\hat{R} = \min\!\left(\frac{u}{d_{sr}},\,1\right),
\label{eq:precision_recall}
\end{equation}
and adopt the $F_{0.5}$ score, which weights precision more heavily than recall, as the core reward signal:
\begin{equation}
F_{0.5} = \frac{1.25 \cdot P \cdot \hat{R}}{0.25 \cdot P + \hat{R}}.
\label{eq:f05}
\end{equation}
To explicitly penalize over-correction, we also track the excess modification ratio
\begin{equation}
\rho = \frac{d_{sp} - d_{sr}}{d_{sr}},\qquad\text{when } d_{sp} > d_{sr}.
\label{eq:rho}
\end{equation}
The full correctness reward $R_{\text{c}}$ is then computed by a piecewise function over five mutually exclusive cases summarized in Table~\ref{tab:reward}.

\begin{table}[t]
  \centering
  \small
  \begin{tabular}{lc}
    \hline
    \textbf{Case} & \textbf{$R_{\text{c}}$} \\
    \hline
    Exact match               & $+3.4$ \\
    Effective \& minimal      & $2.0 \cdot F_{0.5}$ \\
    Effective \& overcorrects & $2.0 \cdot F_{0.5} - 0.8 \cdot \min(\rho,\,2)$ \\
    No effective edit         & $0.0$ \\
    Degrades source           & $-1.5 \cdot \min(-\hat{R},\,1)$ \\
    \hline
  \end{tabular}
  \caption{
    Five branches of the correctness reward $R_{\text{c}}$, clipped to $[-1.5,\,3.4]$. 
    The total reward used during training is $R = R_{\text{f}} + R_{\text{c}}$,
    where $R_{\text{f}} = 0.3$ if the output strictly contains exactly one
    \texttt{<think>} block and one \texttt{<answer>} block, and $0$ otherwise.
  }
  \label{tab:reward}
\end{table}

\subsubsection{The Limitation of GRPO}
\begin{figure}
\small
    \centering
    \includegraphics[width=1\linewidth]{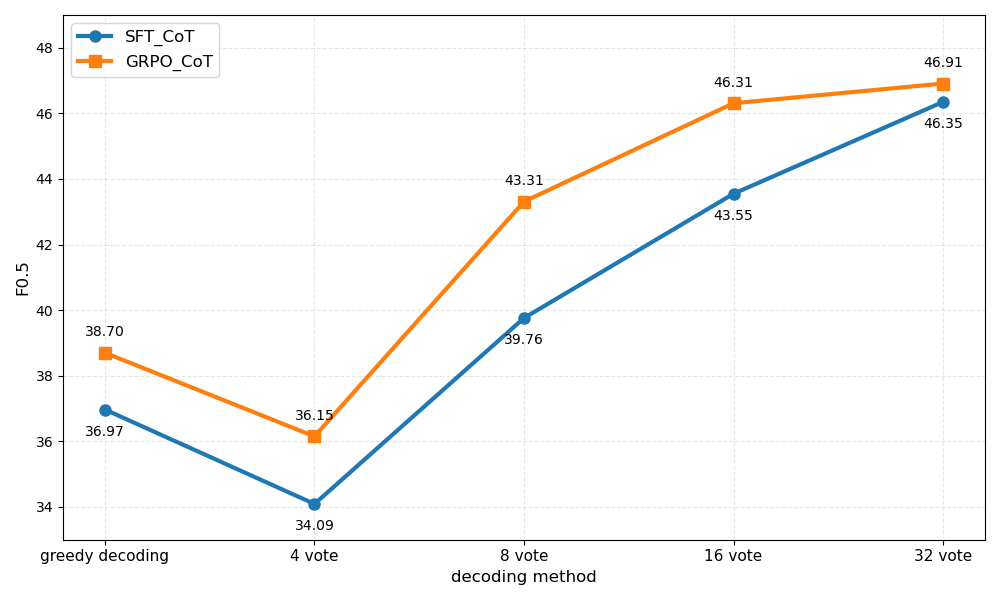}
    \caption{Performance comparison of $F_{0.5}$ scores for SFT and GRPO models across different decoding strategies. SFT\_CoT denotes the model fine-tuned via SFT on CoT data, while GRPO\_CoT refers to the GRPO-optimized model initialized from the SFT checkpoint.}
    \label{fig:compare_SFT_GRPO}
\end{figure}

We first tried GRPO, as shown in Figure \ref{fig:compare_SFT_GRPO}, the model's performance is improved, but the magnitude of improvement remains limited when using 32 vote inference.
We argue that this limitation originates from a fundamental objective mismatch: the optimization objective of standard GRPO is the expected reward of individual samples
\begin{equation}
J_{\text{GRPO}} = \mathbb{E}_{y \sim \pi_{\theta}} \big[ r(y) \big],
\end{equation}
and its advantage is computed through relative comparisons within each group:
\begin{equation}
A_{i,j}^{\text{GRPO}} = \frac{r_{i,j} - \bar{r}_{G_i}}{\sigma_{G_i}}.
\end{equation}
However, the actual self-consistency performance used during inference is determined by
\begin{equation}
E_{G \sim \pi_{\theta}^K}\big[ r(\text{Vote}(G)) \big],
\end{equation}
which is not fully equivalent to the former objective. GRPO indirectly improves vote performance by increasing the mean reward of individual samples. However, the training objective merely improves the overall reward of the rollout group as a whole. It does not specifically push the probability mass of correct answers above that of less-preferred ones, which is precisely what determines whether the best candidate wins under self-consistency.

\subsubsection{Group-Level Relative Policy Optimization}
To align the training process with the objective of improving self-consistency performance, we propose GLPO, a simple yet effective strategy. Building upon the group-normalized advantages in standard GRPO, GLPO introduces a one-sided non-negative vote-margin gain term, which assigns additional positive advantages to samples whose individual rewards exceed the vote reward of their corresponding group, in proportion to the amount by which they exceed it:
\begin{equation}
A_{i,j}^{\text{GLPO}} = 
\underbrace{\frac{r_{i,j} - \bar{r}_{G_i}}{\sigma_{G_i}}}_{\text{base}}
+ \underbrace{\max\bigl(0,r_{i,j}\!-\!R(G_i)\bigr)}_{\text{bonus}},
\end{equation}
where $r_{i,j}$ denotes the individual reward of the $j$-th sample in the $i$-th group, while $\bar{r}_{G_i}$ and $\sigma_{G_i}$ represent the mean and standard deviation of the rewards within that group, respectively, and $R(G_i)$ denotes the vote reward of the group. The pseudocode of the GLPO algorithm is provided in Appendix~\ref{algorithm}.

This design enables the model to establish a gradient allocation mechanism within each group that is more aligned with the self-consistency optimization objective: (1) outputs that outperform the group-level vote score are substantially amplified; (2) answers whose rewards fall below the group vote score remain unaffected and revert to the standard GRPO behavior; and (3) outputs belonging to the vote result are prevented from being excessively reinforced.
Compared with GRPO, which relies solely on relative ranking within groups, GLPO can directly exploit the group preference signals provided by the vote reward, thereby aligning the optimization direction more closely with the ultimate self-consistency objective.

\subsection{Self-Consistency Decoding}\label{nvote}
While greedy decoding selects the single most likely output, our training objective is explicitly aligned with a multi-sample inference procedure. We therefore adopt self-consistency: for each input $x$, the policy $\pi_\theta$ generates $N$ candidate corrections $\{\hat{y}_1, \dots, \hat{y}_N\}$ under stochastic decoding with temperature $T=1.0$, and the final prediction is determined by majority voting:
\begin{equation}
  \hat{y}^{\star} = \arg\max_{y} \sum_{i=1}^{N} \mathbb{1}[\hat{y}_i = y].
\end{equation}

In our experiments, we vary $N \in \{4, 8, 16, 32\}$ to study how voting scale interacts with CoT-induced output diversity, and how it amplifies the gains of GLPO over standard GRPO.

\section{Experiments}

\subsection{Dataset}\label{dataset}
 We evaluate on two datasets, summarized in Table~\ref{tab:dataset}. CSED-C \cite{Sun-etal-2023-gector} is a manually annotated dataset of 10{,}682 sentence pairs collected from Chinese examination correction tasks.
 NaSGEC-Exam is the examination subset of NaSGEC \cite{zhang-etal-2023-nasgec}, containing about 7{,}000 erroneous sentences from real Chinese exams with multiple expert-reviewed reference corrections. More detailed introduction about the two datasets are provided in Appendix~\ref{A_dataset}.

\begin{table}[t]
  \centering
  \small
  \setlength{\tabcolsep}{2pt}
  \begin{tabular}{llrrrr}
    \toprule
    \multirow{2}{*}{\textbf{Dataset}} & \multirow{2}{*}{\textbf{Usage}} &
      \multicolumn{2}{c}{\textbf{Train}} &
      \multirow{2}{*}{\textbf{Val}} & \multirow{2}{*}{\textbf{Test}} \\
    \cmidrule(lr){3-4}
    & & \textbf{Before} & \textbf{After} & & \\
    \midrule
    \multirow{2}{*}{CSED-C}
      & non-CoT & \multirow{2}{*}{8{,}682} & 11{,}338 & \multirow{2}{*}{1{,}000} & \multirow{2}{*}{1{,}000} \\
      & CoT   &                          & 11{,}204 & & \\
    \midrule
    \multirow{2}{*}{NaSGEC-Exam}
      & non-CoT & \multirow{2}{*}{4{,}000} &  5{,}818 & \multirow{2}{*}{1{,}000} & \multirow{2}{*}{2{,}000} \\
      & CoT   &                          &  5{,}406 & & \\
    \bottomrule
  \end{tabular}
  \caption{
    Dataset statistics across training stages.
    Only the training set is processed. `Before' reports the raw sentence count,
    while `After' reports the count after task-specific preprocessing.
  }
  \label{tab:dataset}
\end{table}

\subsection{Baselines}
To evaluate the effectiveness of our method, we select several representative baseline models and methods for comparative experiments. More detailed introduction is presented in Appendix \ref{A_baselines}.

\noindent\textbf{Seq2Seq Models.} We adopt four Seq2Seq baselines: mT5-small and mT5-base \cite{xue-etal-2021-mt5}, two scales of the multilingual pretrained model. BART-large-Chinese \cite{Shao2024}, a denoising autoencoder pretrained specifically for Chinese generation. SynGEC \cite{zhang-etal-2022-syngec}, which augments BART with dependency syntax to better capture complex grammatical errors.

\noindent\textbf{Seq2Edit Models.} We use GECToR \cite{Omelianchuk-etal-2020-gector} as the representative Seq2Edit baseline, which reformulates correction as token-level edit operation prediction (insertion, deletion, substitution) instead of generating the full target sentence.

\noindent\textbf{Large Language Models.} We adopt the LLM-based CSEC method proposed by \citet{WU2026131925}, denoted as CSEC-LLM. This method leverages instruction tuning together with an example selection strategy and a re-scoring mechanism to improve semantic error correction performance.

\subsection{Implementation}\label{implementation}
During the SFT stage, we conduct training based on LLaMA-Factory\footnote{https://github.com/hiyouga/LlamaFactory} and select Qwen3.5-4B as the base model, then adopt the LoRA fine-tuning approach, with the LoRA rank set to 128, batch size set to 4, learning rate set to $2.0\times 10^{-4}$, and the number of training epochs set to 3. Each training run in this stage takes approximately 4 hours.

During the RL stage, we perform our GLPO training based on ms-swift\footnote{https://github.com/modelscope/ms-swift}. LoRA fine-tuning is also adopted in this stage, with the following training configurations: LoRA rank of 8, batch size of 8, learning rate of $1\times 10^{-6}$, rollout number of 8, temperature set to 1.0, $\beta$ set to 0.001, and the number of training epochs set to 5. The training process takes approximately 80 hours. All experiments are conducted on two A30 GPUs.

\begin{table}
  \centering
\resizebox{0.5\textwidth}{!}{
  \begin{tabular}{llllccc}
    \toprule
    \textbf{Dataset} & \textbf{Method} & \textbf{Model} & \textbf{Decoding} & \textbf{P} & \textbf{R} & \textbf{$F_{0.5}$} \\
    \midrule
    \multirow{15}{*}{CSED-C}
      & \multirow{4}{*}{Seq2Seq}
        & mT5-small      & \multirow{4}{*}{--} & 33.70                         & 5.40                           & 16.50 \\
      & & mT5-base       &                     & \underline{\textbf{57.00}}    & 19.00                          & 40.70 \\
      & & BART-large     &                     & 53.80                         & 38.30                          & \underline{\textbf{49.70}} \\
      & & SynGEC         &                     & 53.00                         & 39.50                          & 49.60 \\
      \cmidrule(lr){2-7}
     & \multirow{2}{*}{CSEC-LLM}
        & ChatGLM3-6B      & \multirow{2}{*}{--} & 33.18                         & 27.80                           & 31.94 \\
      & & Baichuan2-7B       &                     & \textbf{35.90}    & 32.95                          & \textbf{35.26} \\

    \cmidrule(lr){2-7}
      & \multirow{7}{*}{Ours}
        & Zero-shot        & greedy   & 11.95 & 10.30 & 11.58 \\
      \cmidrule(lr){3-7}
      &  & \multirow{2}{*}{SFT}
                         & greedy   & 37.28 & 35.80 & 36.97 \\
      & &                & +32 vote & 47.73 & 41.55 & 46.35 \\
      \cmidrule(lr){3-7}
      &  & \multirow{2}{*}{GRPO}
                         & greedy   & 39.76 & 34.98 & 38.70 \\
      & &                & +32 vote & 48.71 & 40.88 & 46.91 \\
    \cmidrule(lr){3-7}
      & & \multirow{2}{*}{VAA-CSEC}
                         & greedy   & 39.19          & 35.50                       & 38.39          \\
      & &                & +32 vote & \textbf{49.34} & \underline{\textbf{42.15}} & \textbf{47.72} \\
    \midrule
    \multirow{14}{*}{\makecell{NaSGEC\\-Exam}}
      & \multirow{1}{*}{Seq2Seq}          & BART-large    & \multirow{1}{*}{--} & \textbf{23.01} & 11.31 & \textbf{19.06} \\
      \cmidrule(lr){2-7}
      &Seq2Edit          & GECToR        &--                     & 20.93          & 8.80  & 16.41 \\
      \cmidrule(lr){2-7}
     & \multirow{2}{*}{CSEC-LLM}
        & ChatGLM3-6B      & \multirow{2}{*}{--} & 30.51                         & 24.30                           & 29.03 \\
      & & Baichuan2-7B       &                     & \textbf{34.46}    & 22.69                          & \textbf{31.22} \\
    \cmidrule(lr){2-7}
      & \multirow{7}{*}{Ours}
        & Zero-shot        & greedy   & 11.92 & 15.10 & 12.44 \\
        \cmidrule(lr){3-7}
      &  & \multirow{2}{*}{SFT}
                         & greedy   & 19.91    & 23.02    & 20.46    \\
      & &                & +32 vote & 47.43    & 26.78    & 41.09    \\
      \cmidrule(lr){3-7}
      &   & \multirow{2}{*}{GRPO}
                         & greedy   & 24.61    & 22.49    & 24.15    \\
      & &                & +32 vote & 46.81    & 26.43    & 40.55    \\
      \cmidrule(lr){3-7}
      &  
        & \multirow{2}{*}{VAA-CSEC}
                         & greedy   & 27.66                       & 24.20 & 26.89 \\
      & &                & +32 vote & \underline{\textbf{48.43}}  & 26.50 & \underline{\textbf{41.55}} \\
    \bottomrule
  \end{tabular}
  }
  \caption{
    Main results on CSED-C and NaSGEC-Exam.
    \underline{\textbf{Underlined bold}} indicates the best score per metric across all methods within each dataset;
    \textbf{bold} indicates the best score per metric among the remaining methods of the same method.
  }
  \label{tab:main_results}
\end{table}

\subsection{Evaluation}
During the evaluation stage, we employ ChERRANT, proposed by \citet{cherrant}, to evaluate the model outputs. This metric is capable of computing precision, recall, and the $F_{0.5}$ score.
The $F_{0.5}$ score is a variant of $F_{1}$ metric that places greater emphasis on the precision of corrections rather than recall. This evaluation criterion is more consistent with the objectives of Chinese semantic error correction, which aim to avoid erroneous modifications and minimize edits. Specifically, the model is expected to make only necessary corrections to erroneous parts rather than extensively rewriting the original sentence, thereby better preserving the semantic meaning and stylistic consistency of the original expression.

\subsection{Main Results}\label{main results}
Table~\ref{tab:main_results} presents the main results on CSED-C and NaSGEC-Exam. We organize our discussion around three findings:

1) Competitive performance with a lightweight model. On CSED-C, our method surpasses larger LLM-based baselines and ranks the best among them, while overall placing second only to BART-large. It achieves the highest recall of 42.15\% across all methods, indicating a stronger ability to identify semantic errors than even the strongest task-specific Seq2Seq baseline. On NaSGEC-Exam, our method outperforms all baselines on both precision and $F_{0.5}$, establishing a new state of the art. The contrast between the two datasets is informative: while traditional Seq2Seq models can still be competitive on the relatively well-defined CSED-C, they struggle considerably on NaSGEC-Exam, whose subtler semantic errors favor LLM-based approaches such as ours.

2) Our VAA-CSEC consistently improves over GRPO and SFT. Across both datasets, VAA-CSEC demonstrates a clear advantage over its GRPO and SFT counterparts. Under 32 Vote decoding, VAA-CSEC reaches 47.72\% $F_{0.5}$ on CSED-C, outperforming GRPO at 46.91\% and SFT at 46.35\%. On NaSGEC-Exam, it achives 41.55\%, again surpassing GRPO at 40.55\% and SFT at 41.09\%. Interestingly, on NaSGEC-Exam, applying standard GRPO on top of SFT yields a slight performance drop of 0.64\%, suggesting that standard group-normalized advantages can degenerate into noisy signals when intra-group reward variance is low. Our GLPO strategy not only recovers the lost performance but surpasses both baselines, indicating that its vote-margin bonus provides a more stable, consistency-based signal that helps the policy escape such suboptimal training dynamics.

3) Self-consistency decoding unlocks substantial gains. Switching from greedy decoding to 32 Vote produces large $F_{0.5}$ improvements across all of our model variants and both datasets. For VAA-CSEC, the absolute gain reaches $+9.33\%$ on CSED-C and $+14.66\%$ on NaSGEC-Exam. This pattern aligns with the conclusion of \citet{ye2022unreliability} that individual CoT reasoning paths can be unreliable: greedy decoding commits to a single, potentially ungrounded trajectory and thus underutilizes the capacity of the learned policy. Aggregating multiple samples via majority voting mitigates this unreliability by stably recovering high-quality corrections to which the policy has already assigned non-trivial probability mass. Furthermore, as shown in the table \ref{tab:cost_performance}, while the model performs best with 32 votes, the inference cost increases exponentially. Considering inference efficiency, we recommend using a 16-vote setting in practical applications.

\begin{table}[t]
\centering
\resizebox{0.5\textwidth}{!}{
\begin{tabular}{lccccc}
\toprule
\textbf{VAA-CSEC} & \textbf{P} & \textbf{R} & $\mathbf{F_{0.5}}$ & \textbf{Time cost} & $\mathbf{\Delta F_{0.5}}$ \\
\midrule
Greedy & 39.19 & 35.50 & 38.39 & $\sim$2 min  & --    \\
4-vote          & 39.51 & 36.02 & 38.76 & $\sim$8 min  & +0.37 \\
8-vote          & 42.00 & 37.67 & 41.06 & $\sim$16 min & +2.30 \\
16-vote         & 47.55 & 40.58 & 45.97 & $\sim$32 min & +4.91 \\
32-vote         & 49.34 & 42.15 & 47.72 & $\sim$64 min & +1.75 \\
\bottomrule
\end{tabular}
}
\caption{
Cost-performance trade-off of VAA-CSEC on CSED-C.
}
\label{tab:cost_performance}
\end{table}

\subsection{Ablation Study}
\label{sec:ablation}
To understand the contribution of each component, we conduct an ablation study of CoT distillation and GLPO on CSED-C, as shown in Table~\ref{tab:ablation study}.

Removing CoT distillation causes the largest performance drop of $4.49\%$, indicating that the structured reasoning trajectories distilled from the teacher are critical for establishing a strong initial policy. Replacing GLPO with vanilla GRPO leads to a smaller degradation of $0.81\%$. The decline is mainly reflected in recall, which decreases by $1.27\%$, while precision drops by only $0.63\%$. This result suggests that the vote-margin signal introduced by GLPO helps the policy recover a larger proportion of true corrections.

Removing both components yields an $F_{0.5}$ score of 43.64\%, which is slightly higher than the score of 43.23\% obtained when removing only CoT distillation. We attribute this phenomenon to the lack of rollout diversity in the absence of CoT training. Even with a sampling temperature of 1.0, the policy generates nearly identical trajectories, causing the intra-group reward variance to collapse. As a result, neither GRPO nor GLPO can provide informative optimization signals. The resulting difference of 0.41\% therefore falls within expected training noise and further supports that GLPO is most effective when applied to CoT-trained policies.
\begin{table}[t]
    \small
    \centering
    \begin{tabular}{cccc}
        \toprule
        Method  & P & R & $F_{0.5}$ \\
        \midrule
        VAA-CSEC & \textbf{49.34} & \textbf{42.15} & \textbf{47.72} \\
        $w$/$o$ CoT  & 44.27 & 39.54 & 43.23 \\
        $w$/$o$ GLPO  & 48.71 & 40.88 & 46.91 \\
        $w$/$o$ CoT and GLPO  & 44.66 & 39.99 & 43.64 \\
        \bottomrule
    \end{tabular}
    \caption{Ablation study of the proposed components on the CSEC task. ``$w$/$o$'' denotes removing the corresponding module from the full model.}
    \label{tab:ablation study}
\end{table}


\subsection{Cross-Domain Evaluation}
\label{sec:cross_domain}
We make the following observations from Table~\ref{tab:cross_domain}.
The two training sets induce noticeably different decision behaviors. Models trained on CSED-C consistently achieve higher recall on both test sets: under 32 Vote, the CSED-C-trained model reaches 42.15\% and 42.14\% Recall on the two test sets, substantially higher than the 27.88\% and 26.50\% Recall obtained by the model trained on NaSGEC-Exam. Conversely, training on NaSGEC-Exam yields higher precision: 50.54\% and 48.43\% Precision versus 49.34\% and 35.22\% Precision from the CSED-C trained counterpart. This contrast suggests that the two corpora encode distinct error distributions and labeling tendencies. CSED-C, with its broader coverage of semantic error types, drives the policy toward a more aggressive editing behavior, whereas the more conservative NaSGEC-Exam annotations push the policy toward editing only when strongly confident.

 Despite reasonable cross-domain transfer, the highest $F_{0.5}$ on each test set is still obtained by the in-domain model: 47.72\% on CSED-C and 41.55\% on NaSGEC-Exam, both under 32 Vote decoding. This confirms that our method is able to fit each dataset's specific error distribution, while still retaining enough generalization for usable out-of-domain performance.

\begin{table}[t]
\centering
\resizebox{0.5\textwidth}{!}{
\begin{tabular}{lllccc}
\toprule
\textbf{Train Set} & \textbf{Test Set} & \textbf{Decoding} & P & R & $F_{0.5}$\\ 
\midrule
\multirow{4}{*}{CSED-C}   & \multirow{2}{*}{CSED-C}     & greedy   & 39.19          & 35.50                       & 38.39\\
&         & +32 Vote    & 49.34 & \textbf{42.15} & \textbf{47.72}\\
\cmidrule(lr){2-6}
 & \multirow{2}{*}{NaSGEC-Exam} & Greedy & 27.06 & 36.46 & 28.53 \\
&         & +32 Vote    & 35.22   & \textbf{42.14}   & 36.42 \\
\midrule
\multirow{4}{*}{NaSGEC-Exam}   & \multirow{2}{*}{CSED-C}     & Greedy     & 35.65 & 26.46 & 33.33\\
&         & +32 Vote    & \textbf{50.54} & 27.88 & 43.47\\
\cmidrule(lr){2-6}
 & \multirow{2}{*}{NaSGEC-Exam} & Greedy & 27.66   & 24.20   & 26.89 \\
&         & +32 Vote    & \textbf{48.43}   & 26.50   & \textbf{41.55} \\
\bottomrule
\end{tabular}
}
\caption{
  Cross-domain evaluation results.
  Each model is trained on one dataset and evaluated on both test sets.
}
\label{tab:cross_domain}
\end{table}

\section{Conclusion}
We presented VAA-CSEC, a multi-stage framework for CSEC that targets two obstacles to applying LLMs to this task. The first is 
over-correction, and the second is the unclear interaction between 
CoT reasoning and sampling-based decoding. To address the first, we designed a task-specific reward that combines format constraints, edit-distance correctness, and an explicit over-correction penalty, directly aligning the training objective with the minimal-editing principle of CSEC. To address the second, we introduced GLPO, a group-level extension of GRPO that reallocates advantages based on the margin between each rollout's reward and a vote-aggregated group reward, making the RL objective at training time consistent with the self-consistency objective used at inference time. VAA-CSEC outperforms all LLM-based baselines on CSED-C and establishes a new state of the art on NaSGEC-Exam, with consistent improvements over both SFT and GRPO and substantial additional gains from self-consistency decoding.

Beyond the empirical results, our work also reinforces a broader 
methodological observation already noted in prior studies on reasoning with LLMs. The benefit of CoT is conditional on the sampling strategy, and our preliminary analysis on CSEC provides further evidence for this view. Under greedy decoding CoT can even hurt, while under self-consistency it yields steadily growing gains that eventually surpass the non-CoT baseline. This decoupling between training-time and inference-time objectives is not specific to CSEC, and we expect that the GLPO recipe of using an in-group voting consensus as an explicit optimization anchor will be applicable to other generation tasks that rely on self-consistency decoding.

\section*{Limitations}
While VAA-CSEC achieves strong empirical results on Chinese semantic error correction, several aspects of our study are constrained by available compute and the scope of our experimental design, and we discuss them below. Due to computational constraints, our parameter study only covers 
rollout sizes $N \in \{4, 6, 8\}$. Whether the observed trend continues at larger scales, and in particular whether GRPO and VAA-CSEC behave differently as $N$ grows beyond 8, remains an open question. The potential interaction between the training-time rollout size and the test-time self-consistency scale also warrants further investigation. Furthermore, our experiments focus exclusively on CSEC, evaluated on CSED-C and 
NaSGEC-Exam. While the underlying mechanism of GLPO is language-agnostic and applies to any task where self-consistency aggregates over multiple samples, its empirical applicability to other languages, particularly those with substantially different morphology or error patterns, has not been validated.

\section*{Ethics Statement}
The datasets and large language models used in our study come from open-access repositories. This ensures that we comply with all relevant ethical standards and authorizations. We strictly follow established research ethics throughout our research. As for the AI assistant, we utilize ChatGPT and Claude to identify textual errors and polish paper.

\section*{Acknowledgments}

This work was supported by Guangdong Basic and Applied Basic Research Foundation (No. 2025A1515110214).

\bibliography{ref}

\clearpage
\appendix

\begin{figure*}[t]
  \centering
  \includegraphics[width=1\textwidth]{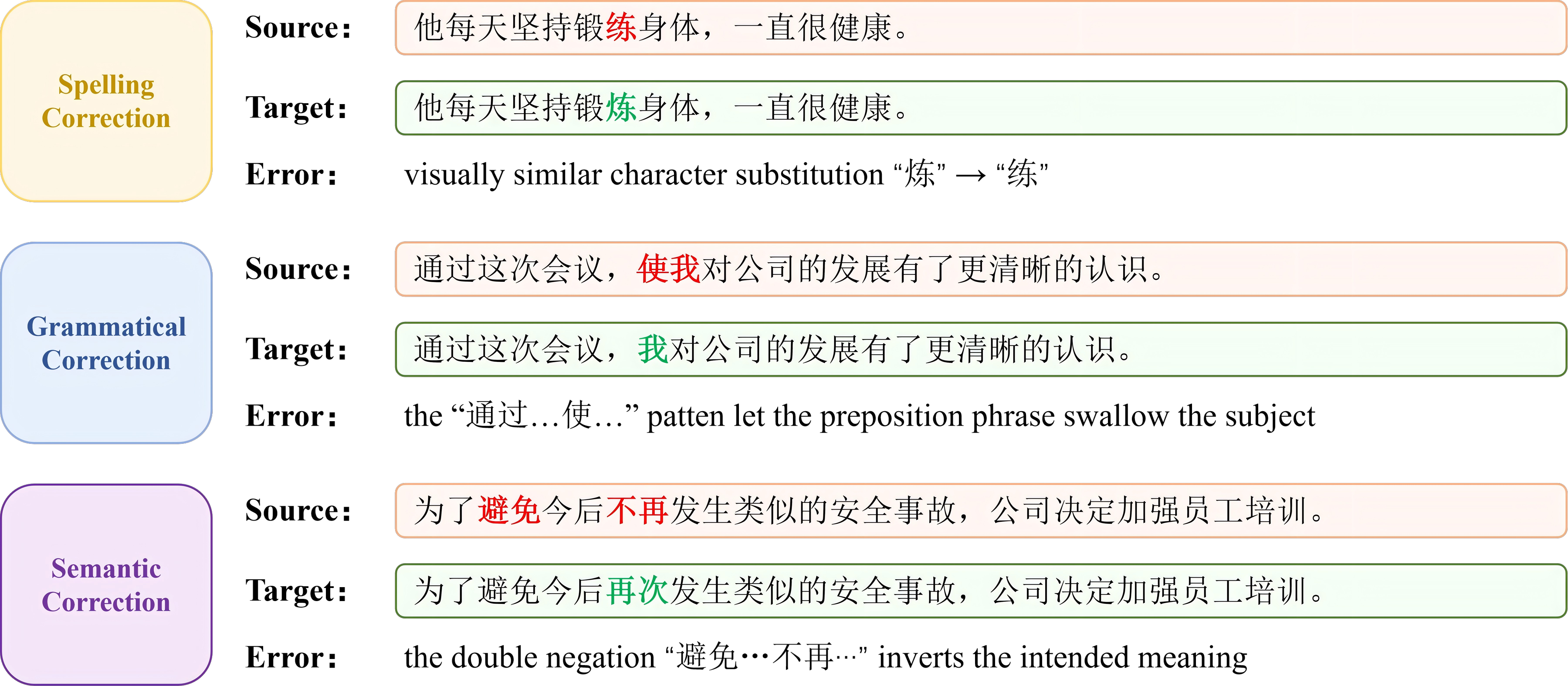}
  \caption{
    Three categories of Chinese text error correction, illustrating a progression
    from surface-level character errors to deep semantic errors.
    Errors in the source sentences are highlighted in red, and the corresponding
    edits in the target sentences are shown in green.
  }
  \label{fig:examples}
\end{figure*}

\section{Detailed Examples of Different Tasks}\label{example}

Spelling error correction targets surface-level character errors such as typographical mistakes, visually similar characters, and incorrect pinyin spellings. Grammatical error correction addresses syntactic well-formedness, including improper word order and inappropriate collocations. Semantic error correction goes further, targeting logical relations, semantic compatibility, and contextual coherence. Even sentences free of spelling and grammatical errors may still require revision when their meaning is incoherent. Figure \ref{fig:examples} illustrates the three categories.

\section{Template}\label{A_template}
\begin{figure*}[htbp]
    \centering
    \makebox[\linewidth]{%
        \includegraphics[width=\linewidth]{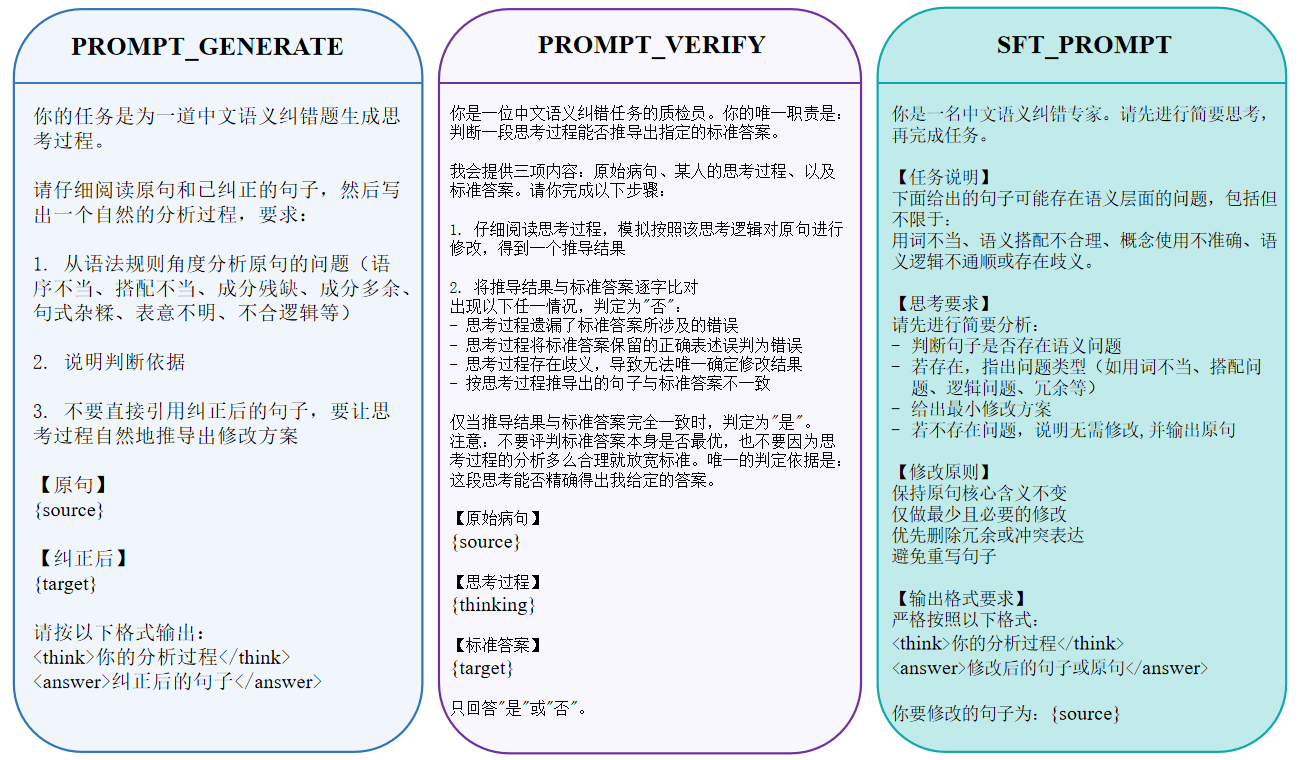}%
    }
    \caption{Template for Data Processing}
    \label{fig:prompt}
\end{figure*}

The detailed template for data processing is shown in Figure~\ref{fig:prompt}. The English translation of the templates is as follows.

\begin{lstlisting}[style=promptstyle, caption={Prompt template for generating chain-of-thought reasoning during CoT distillation.}, label={lst:prompt_generate}]
# Task
Your task is to generate a reasoning process for a Chinese semantic error correction problem.
Please carefully read the original sentence and the corrected sentence, then write a natural analysis process with the following requirements:
1. Analyze the problems in the original sentence from the perspective of grammatical rules, such as incorrect word order, improper collocation, missing components, redundant components, mixed sentence patterns, ambiguous expression, illogical meaning, etc.
2. Explain the basis for your judgment.
3. Do not directly quote the corrected sentence. Instead, let the reasoning process naturally lead to the correction.
# Input
[Original Sentence]: {source}
[Corrected Sentence]: {{target}}
# Output Format
<think>Your reasoning process</think>
<answer>The corrected sentence</answer>
\end{lstlisting}

\begin{lstlisting}[style=promptstyle, caption={Prompt template for verifying the quality of distilled reasoning trajectories.}, label={lst:prompt_verify}]
# Role
You are a quality inspector for a Chinese semantic error correction task. Your only responsibility is to determine whether a given reasoning process can uniquely derive the specified reference answer.
# Procedure
I will provide three items: the original erroneous sentence, a reasoning process written by someone, and the reference answer. Please complete the following steps:
1. Carefully read the reasoning process and simulate modifying the original sentence according to that reasoning to obtain a derived result.
2. Compare the derived result with the reference answer character by character.
3. If any of the following conditions occur, the judgment must be "No":
   - The reasoning process misses an error that is corrected in the reference answer.
   - The reasoning process incorrectly identifies a correct expression retained in the reference answer as erroneous.
   - The reasoning process is ambiguous, making it impossible to uniquely determine the final corrected sentence.
   - The sentence derived from the reasoning process is inconsistent with the reference answer.
4. Only when the derived result is completely identical to the reference answer should the judgment be "Yes".
# Note
Do not evaluate whether the reference answer itself is optimal, and do not relax the standard simply because the reasoning process appears reasonable. The only criterion is whether the reasoning process can precisely derive the given reference answer.
# Input
[Original Sentence]: {source}
[Reasoning Process]: {thinking}
[Reference Answer]: {target}
# Output
Reply only with "Yes" or "No".
\end{lstlisting}

\begin{lstlisting}[style=promptstyle, caption={Prompt template used at SFT training and inference time.}, label={lst:prompt_sft}]
# Role
You are an expert in Chinese semantic error correction. Please first conduct brief reasoning, then complete the task.
# Task Description
The following sentence may contain semantic-level issues, including but not limited to:
- inappropriate word usage,
- unreasonable semantic collocations,
- inaccurate concept usage,
- illogical semantic flow,
- ambiguity.
# Reasoning Requirements
Please first provide a brief analysis:
- Determine whether the sentence contains semantic problems.
- If problems exist, identify the type of issue (e.g., inappropriate wording, collocation issue, logical issue, redundancy, etc.).
- Provide the minimal modification needed.
- If no problem exists, explain that no modification is necessary and output the original sentence.
# Modification Principles
- Preserve the core meaning of the original sentence.
- Make only the minimal and necessary modifications.
- Prefer deleting redundant or conflicting expressions.
- Avoid rewriting the entire sentence.
# Output Format
Strictly follow the format below:
<think>Your reasoning process</think>
<answer>The corrected sentence or the original sentence</answer>
# Input
[Sentence to revise]: {source}
\end{lstlisting}

\section{Dataset}\label{A_dataset}
We conduct our experiments on two datasets: CSED-C and NaSGEC-Exam. CSED-C is a high-quality manually annotated dataset designed for deep semantic error correction in Chinese. It is primarily collected from sentence correction questions in middle and high school Chinese language examinations and contains a total of 10,682 sentence pairs. The dataset covers various complex semantic and grammatical error types, including incorrect word order, collocation errors, missing sentence components, mixed sentence structures, and logical inconsistencies. Compared with traditional grammatical error correction datasets, the errors in CSED-C are more subtle and the sentence expressions are more natural, thereby imposing higher demands on the model’s semantic understanding and reasoning capabilities. Consequently, it has become an important benchmark dataset for research on deep semantic error correction in Chinese. 

NaSGEC-Exam is a Chinese language examination subset of the NaSGEC dataset, containing approximately 7,000 erroneous Chinese sentences and their corresponding corrections collected from real Chinese language examination sentence correction tasks. The dataset is constructed through independent annotation and expert review, providing multiple high-quality reference answers for each sample. It covers various typical semantic and grammatical errors, including improper collocations, missing sentence components, word order errors, mixed sentence structures, and logical inconsistencies. As it effectively reflects the real-world distribution of erroneous sentences in native Chinese contexts, it has been widely used in research on Chinese semantic error correction, multi-reference answer modeling, and cross-domain generalization.

\section{Baselines}\label{A_baselines}
\noindent\textbf{Seq2Seq Models.} The Seq2Seq baselines adopted in our experiments include mT5-small, mT5-base, BART-large-Chinese, and SynGEC. mT5 \cite{xue-etal-2021-mt5} is a multilingual pretrained model extended from the T5 framework. It adopts a unified text-to-text paradigm and can be applied to various natural language processing tasks, including machine translation, text generation, and grammatical error correction.
The primary differences between mT5-small and mT5-base lie in their parameter scale and representation capability. BART-large-Chinese \cite{Shao2024} is a pretrained generative model specifically developed for Chinese scenarios. It is trained with a denoising autoencoding objective and demonstrates strong performance in Chinese text generation and correction tasks. SynGEC \cite{zhang-etal-2022-syngec} further incorporates syntactic structural information into the BART architecture. By integrating dependency syntax knowledge, it enhances the model’s capability to capture complex grammatical errors and achieves strong performance in Chinese grammatical error correction tasks.

\noindent\textbf{Seq2Edit Models.} Among Seq2Edit approaches, we select GECToR \cite{Omelianchuk-etal-2020-gector} as the representative model. Unlike traditional Seq2Seq methods that directly generate target sentences, GECToR formulates the correction task as edit operation prediction, progressively revising the original sentence through operations such as insertion, deletion, and substitution. The model is typically pretrained on large-scale datasets and subsequently fine-tuned on manually annotated correction corpora, thereby improving its capability to detect and correct grammatical errors while maintaining inference efficiency.

\noindent\textbf{Large Language Models.} We adopt the LLM-based Chinese Semantic Error Correction framework proposed by \citet{WU2026131925}, referred to as CSEC-LLM. This method explores the capability of LLMs for Chinese Semantic Error Correction through instruction tuning, supplemented by a plug-and-play example selection strategy and a re-scoring mechanism. Specifically, the example selection strategy captures fine-grained semantic information from multiple dimensions to retrieve high-quality demonstrations, while the re-scoring mechanism selects the optimal correction from multiple candidates generated by LLMs. Experimental results on the CSED-C and NaSGEC-Exam datasets demonstrate that CSEC-LLM substantially improves correction performance, highlighting the effectiveness of LLMs for semantic-level Chinese text correction.

\section{Training Details in RL}\label{GRPO_detail}

\begin{figure*}[htbp]
    \centering
    \begin{minipage}{0.32\textwidth}
        \centering
        \includegraphics[width=\linewidth]{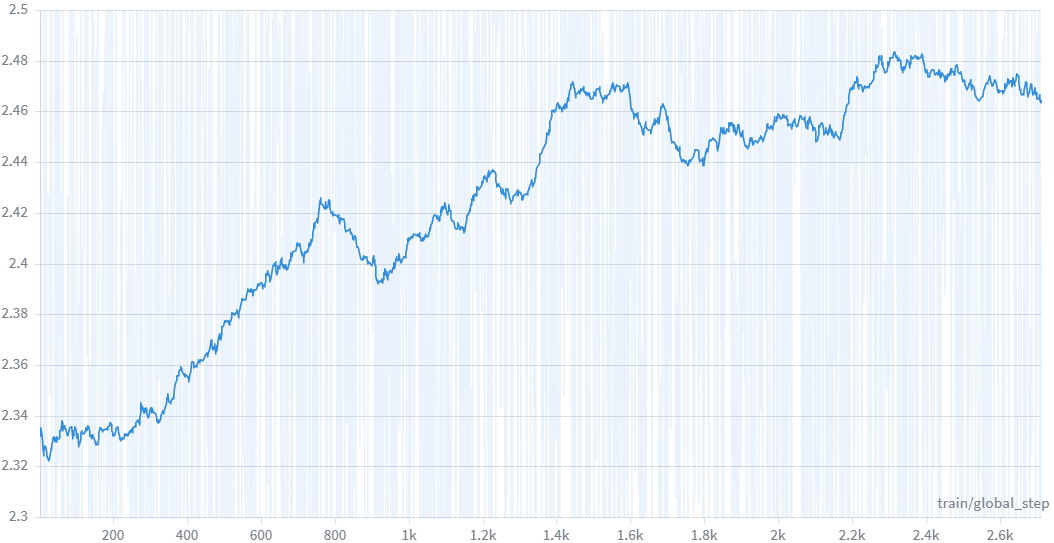}
        \\ (a) Reward
    \end{minipage}
    \hfill
    \begin{minipage}{0.32\textwidth}
        \centering
        \includegraphics[width=\linewidth]{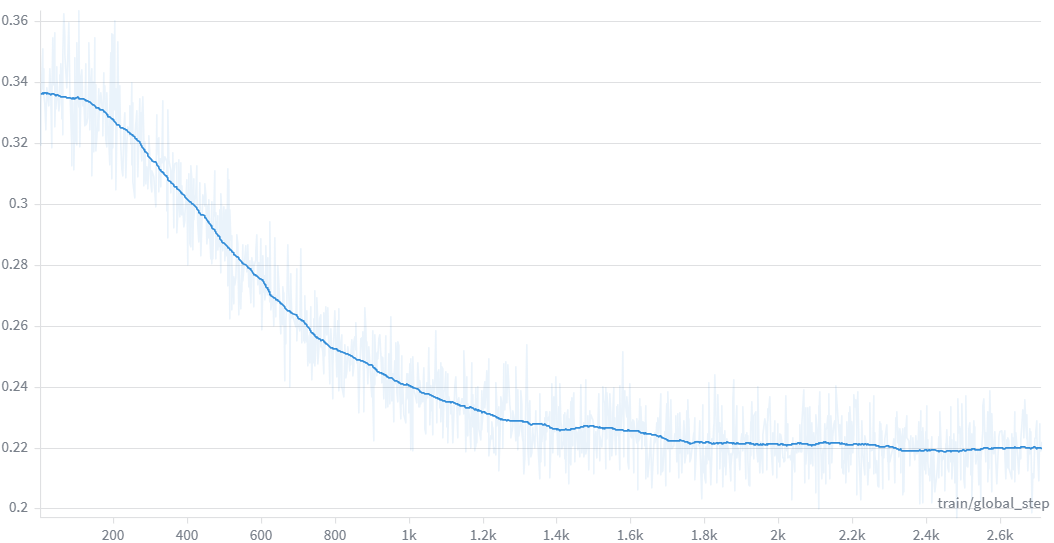}
        \\ (b) Policy Entropy
    \end{minipage}
    \hfill
    \begin{minipage}{0.32\textwidth}
        \centering
        \includegraphics[width=\linewidth]{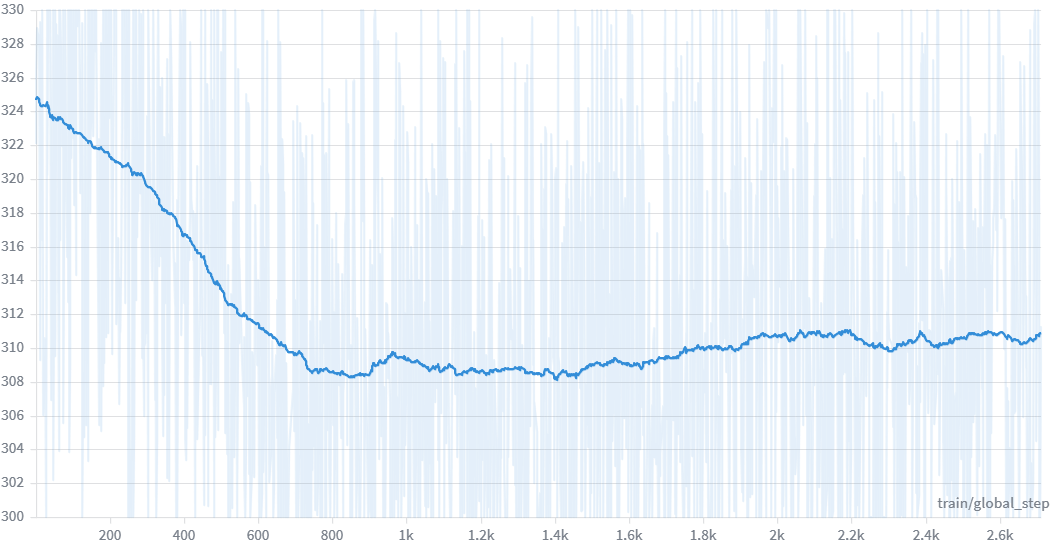}
        \\ (c) Generated Sequence Length
    \end{minipage}
    \caption{Training curves of GLPO during reinforcement learning: (a) reward, (b) policy entropy, and (c) generated sequence length.}
    \label{fig:glpo_training}
\end{figure*}

\begin{figure*}[htbp]
    \centering
    \begin{minipage}{0.32\textwidth}
        \centering
        \includegraphics[width=\linewidth]{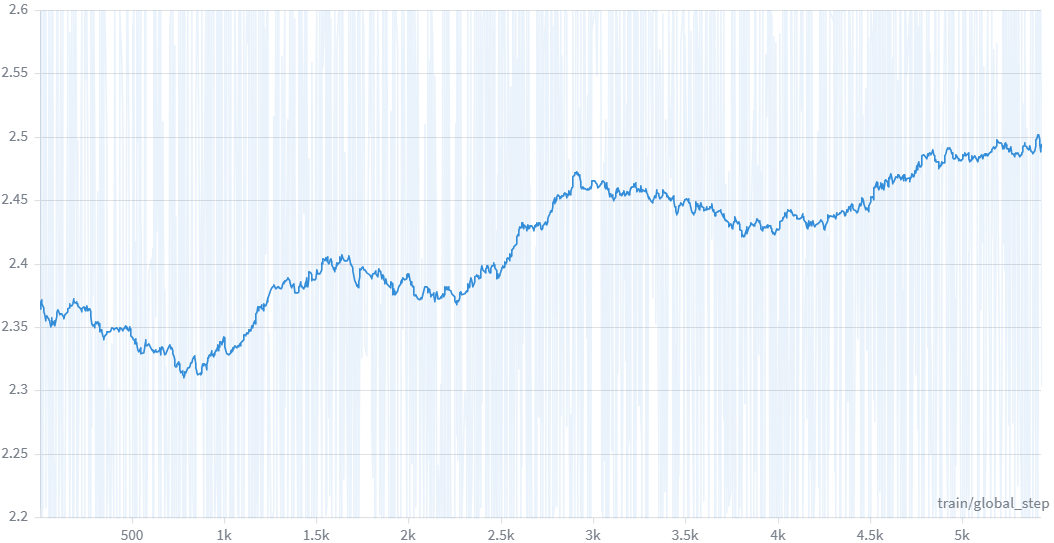}
        \\ (a) Reward
    \end{minipage}
    \hfill
    \begin{minipage}{0.32\textwidth}
        \centering
        \includegraphics[width=\linewidth]{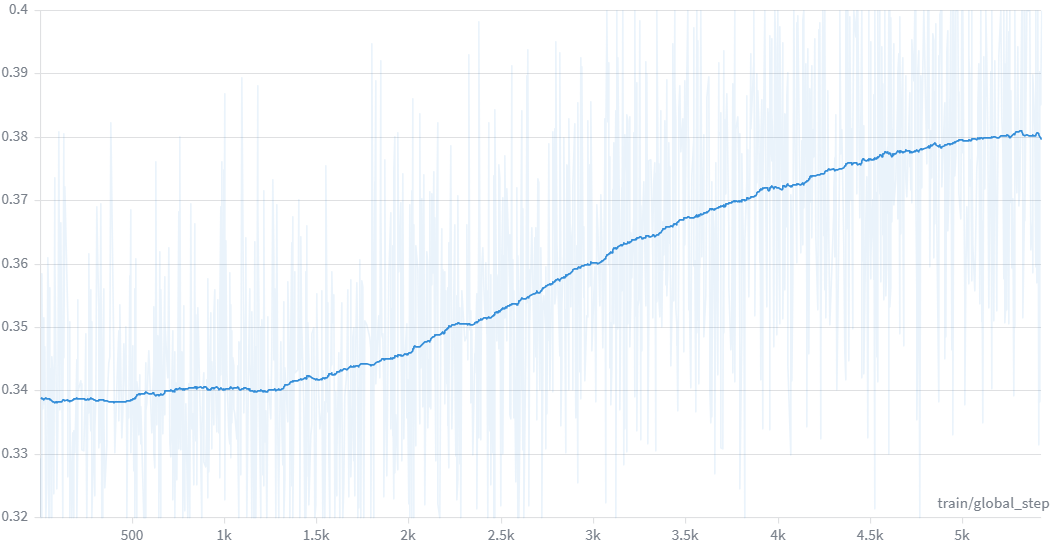}
        \\ (b) Policy Entropy
    \end{minipage}
    \hfill
    \begin{minipage}{0.32\textwidth}
        \centering
        \includegraphics[width=\linewidth]{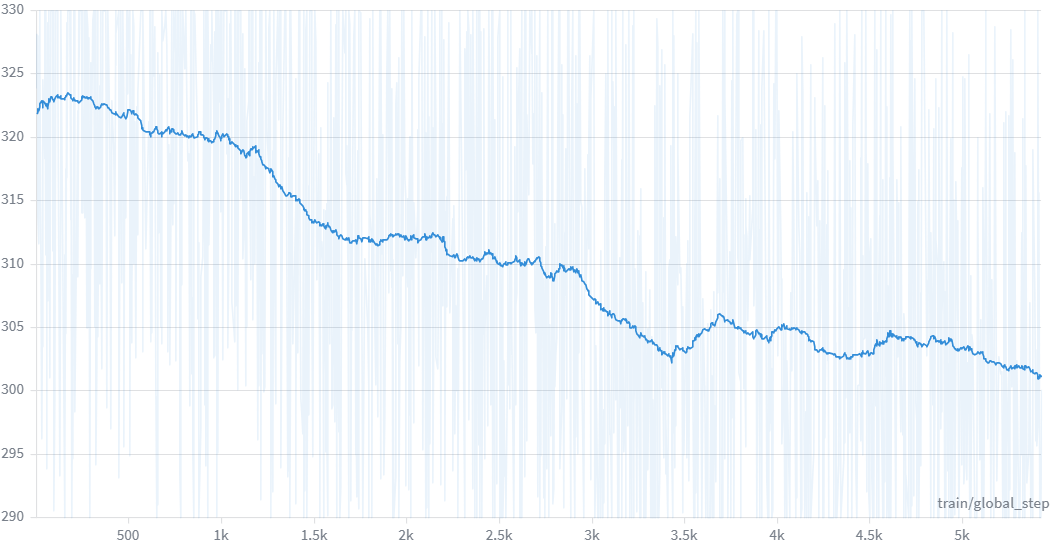}
        \\ (c) Generated Sequence Length
    \end{minipage}
    \caption{Training curves of GRPO during reinforcement learning: (a) reward, (b) policy entropy, and (c) generated sequence length.}
    \label{fig:grpo_training}
\end{figure*}

Figures~\ref{fig:glpo_training} and~\ref{fig:grpo_training} report the training curves of GLPO and GRPO under identical settings, tracking the mean group reward, the policy entropy, and the mean length of generated sequences.

For GLPO, the reward rises rapidly within the first $\sim$1k steps and then begins to oscillate. This is because GLPO does not update the policy toward the highest average reward, but toward the intra-group voting reward. In contrast, the policy entropy and the mean generation length both drop quickly and then stabilize, indicating that the model has converged to a more deterministic editing strategy under GLPO training.

For GRPO, the reward rises steadily and slowly throughout training, which matches our expectation and shows that the reward function correctly guides the policy along the intended optimization direction. Opposite to GLPO, the policy entropy keeps increasing, indicating that the model is still actively exploring new correction strategies, while the mean generation length continues to decrease.

The opposite entropy trends are a direct consequence of the two algorithms' objectives: GRPO's relative-ranking advantage provides little gradient signal when candidates are of similar quality, which is common in CGEC, where multiple valid edits often coexist, so the policy keeps exploring and entropy rises. GLPO's voting anchor, by contrast, amplifies gradients on the consensus trajectory and drives the policy to concentrate on it, while leaving enough residual stochasticity for inference-time voting to remain meaningful.

\section{Case Study}
\label{sec:case_study}
\begin{figure*}
    \centering
    \includegraphics[width=1\linewidth]{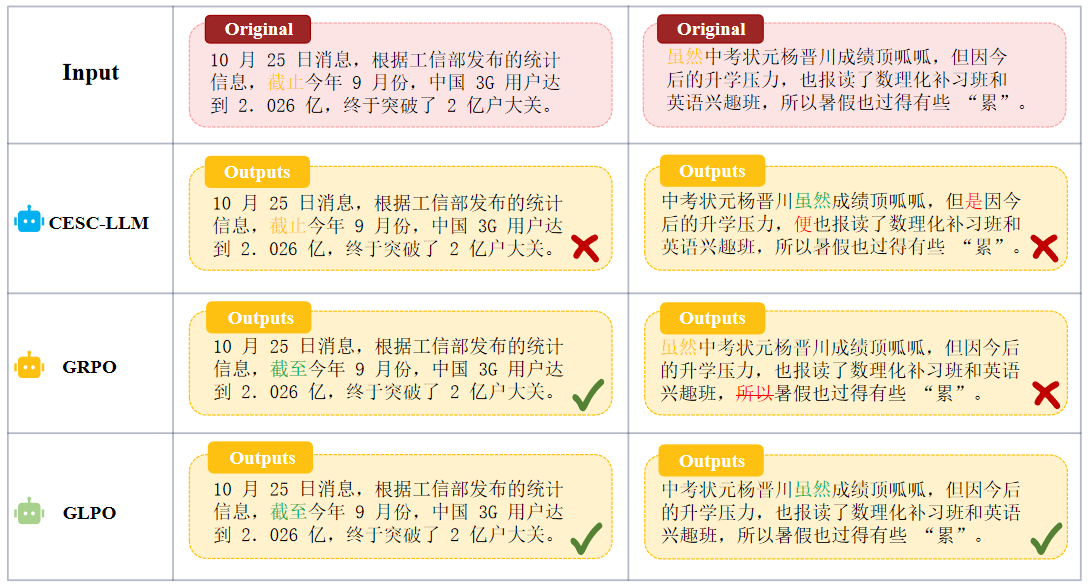}
    \caption{Two examples of case study}
    \label{fig:case_study}
\end{figure*}
To complement the quantitative results, we present two representative cases in Figure~\ref{fig:case_study}.

The first case contains a common confusion between two near-homophones: \textit{jiezhi-1} (to stop'', a verb) and \textit{jiezhi-2} (up to'', a temporal preposition). The supervised baseline CSEC-LLM leaves the sentence unchanged, suggesting that pure SFT is insufficient to capture such fine-grained distinctions. In contrast, both GRPO and our GLPO correctly substitute the verb form with the prepositional form, indicating that reinforcement learning with an edit-aware reward effectively pushes the model toward identifying and correcting such localized errors.

The second case is harder: the original sentence opens with the concessive marker although'' (\textit{suiran}), but its overall structure follows a causal pattern (because\dots so\dots''), making the concessive marker logically inconsistent, which should simply be deleted. This is a typical semantic error that is locally well-formed yet globally incoherent. CSEC-LLM attempts a surface-level rewrite that does not address the underlying logical conflict. GRPO recognizes that an edit is required, but applies it in the wrong place: it preserves the concessive marker and instead removes the causal connective, breaking the causal chain. Only GLPO performs the correct minimal edit by removing the redundant concessive marker, restoring logical consistency while leaving the rest of the sentence intact.

These two cases illustrate the progressive advantage of our framework: RL training enables the model to recover from simple but easily-missed errors, while GLPO's group-level advantage allocation further encourages the model to choose the correct edit when multiple plausible revisions exist, identifying the minimum-edit repair among multiple valid candidates, which we have argued is the central challenge of Chinese semantic correction.

\section{Exploration of Data Ratio}

\begin{table}[t]
    \centering
    \small
    \setlength{\tabcolsep}{4pt}
    \renewcommand{\arraystretch}{1.05}
    \begin{tabular}{@{}llcccc@{}}
        \toprule
        \textbf{Ratio} & \textbf{Mode} & \textbf{Decoding} & \textbf{P} & \textbf{R} & \textbf{$F_{0.5}$} \\
        \midrule
        \multirow{2}{*}{non-CoT} 
            & \multirow{2}{*}{$w/o$ thinking} & Greedy    & 48.03 & 37.37 & 45.44 \\
            &                             & +32 vote  & 48.07 & 38.19 & 45.71 \\
        \midrule
        \multirow{2}{*}{CoT} 
            & \multirow{2}{*}{$w/$ thinking}   & Greedy    & 37.28 & 35.80 & 36.97 \\
            &                             & +32 vote  & 47.73 & 41.55 & 46.35 \\
        \midrule
        \multirow{4}{*}{1:1 mix} 
            & \multirow{2}{*}{$w/o$ thinking} & Greedy    & 44.27 & 38.94 & 43.09 \\
            &                             & +32 vote  & 45.48    & 40.66    & 44.43    \\
            \cmidrule(lr){2-6}
            & \multirow{2}{*}{$w/$ thinking}   & Greedy    & 36.11 & 33.03 & 35.45 \\
            &                             & +32 vote  & \textbf{48.81} & \textbf{41.33} & \textbf{47.10} \\
        \bottomrule
    \end{tabular}
    \caption{Performance comparison of models trained with different CoT to non-CoT data ratios.}
    \label{tab:data_ratio}
\end{table}

As mentioned in Figure \ref{fig:compare}, models trained with CoT data and models trained without CoT data exhibit complementary characteristics. The former performs poorly under greedy decoding at test time, but can obtain substantial gains through N vote sampling. The latter performs strongly under greedy decoding, yet benefits little from majority voting. This observation raises a natural question: can we obtain a unified model that inherits the strengths of both types by mixing CoT and non-CoT data in the SFT training set? Ideally, such a model would possess both ''think'' and ''no-think'' capabilities, so that it maintains strong performance under greedy decoding while also gaining further improvements through N vote.

To this end, we train models under different ratios of CoT to non-CoT data, with the instruction explicitly specifying whether a chain of thought should be produced. The experimental results are shown in Table~\ref{tab:data_ratio}.

We find that when the SFT training data consists of CoT and non-CoT samples mixed in a 1:1 ratio, the model faithfully follows the instruction regarding whether to output the reasoning process. More importantly, the behavior of this mixed model aligns well with our expectations. When instructed to produce a chain of thought, its performance under both greedy decoding and N vote surpasses that of the model trained solely on CoT data. This advantage becomes especially pronounced at $N=32$, where it clearly outperforms both the CoT-only and non-CoT only baselines. When instructed to answer directly, the model still retains the strong greedy decoding performance characteristic of non-CoT training. These results indicate that mixing CoT and non-CoT data is not merely a compromise between the two training schemes, but rather yields a model that is stronger along both dimensions simultaneously.

\section{Parameter Exploration}

The number of rollouts $N$ directly controls the size of the candidate group from which GLPO computes its group-normalized advantage and vote-based bonus. A small $N$ yields fewer comparisons within each group and may underestimate the group-level reward, while a large $N$ increases computational cost roughly linearly. To find a reasonable balance, we vary $N \in \{4, 6, 8\}$ and report results in Table~\ref{tab:variations}.

\begin{table}[t]
\centering
\small
\begin{tabular}{ccccc}
\toprule
$N$  & P & R& $F_{0.5}$ & Agree. \\
\midrule
4  & 48.89 & \textbf{42.68} & 47.50  & 47.6\% \\
6  & 47.88 & 40.51 & 46.20  & 49.9\% \\
8  & \textbf{49.34} & 42.15 & \textbf{47.72} & 47.7 \%   \\
\bottomrule
\end{tabular}
\caption{Parameter study: effect of the number of rollouts $N$ on CSED-C. ``Agree.'' reports the average agreement rate of 32-vote inference using the corresponding trained policy.}
\label{tab:variations}
\end{table}

\begin{algorithm}[htpb]
\small
\caption{GLPO: Group-Level Policy Optimization}
\label{alg:glpo}
\SetKwInOut{Input}{Input}\SetKwInOut{Output}{Output}
\Input{Policy $\pi_\theta$, prompt $x$, individual reward $R(\cdot)$, group size $K$, 
       bonus scale $\alpha$, std floor $\epsilon$}
\Output{Advantages $\{A_1, \ldots, A_K\}$ for policy update}
\BlankLine

\tcc{Step 1: Sample $K$ rollouts and compute per-sample rewards}
$\{y_1, y_2, \ldots, y_K\} \sim \pi_\theta(\cdot \mid x)$\;
$r_i \leftarrow R(y_i, x)$ \textbf{for} $i = 1, \ldots, K$\;
\BlankLine

\tcc{Step 2: Compute group voting reward (constant across the group)}
$y^\star \leftarrow \mathrm{MajorityVote}(\{y_1, \ldots, y_K\})$\;
$R_G \leftarrow R(y^\star, x)$\;
\BlankLine

\tcc{Step 3: Compute GRPO-style base advantage via in-group standardization}
$\mu \leftarrow \frac{1}{K}\sum_{i=1}^{K} r_i$\;
$\sigma \leftarrow \max\!\left(\sqrt{\tfrac{1}{K-1}\sum_{i=1}^{K}(r_i - \mu)^2},\ \epsilon\right)$\;
$A_i^{\text{base}} \leftarrow (r_i - \mu) / \sigma$ \textbf{for} $i = 1, \ldots, K$\;
\BlankLine

\tcc{Step 4: Add vote-margin bonus for samples that surpass the group leader}
\For{$i = 1, \ldots, K$}{
  $\text{bonus}_i \leftarrow \alpha \cdot \max(0,\ r_i - R_G)$\;
  $A_i \leftarrow A_i^{\text{base}} + \text{bonus}_i$\;
}
\BlankLine

\Return{$\{A_1, \ldots, A_K\}$}
\end{algorithm}

We observe that the model's performance is sensitive to $N$ in a non-monotonic manner: with $N=4$, our model already attains a competitive $F_{0.5}$ of 47.5\%, performance drops to 46.2\% at $N=6$, and rises again to its peak of 47.72\% at $N=8$. We attribute the dip at $N=6$ to insufficient candidate diversity: six samples are not enough to consistently produce a clear majority for vote-based aggregation, yet are too few to give a stable group baseline. With $N=8$, the candidate group is large enough to produce a confident vote reward and to sharpen the policy through GLPO's vote-margin bonus, leading to the strongest result among all tested configurations. In addition, we can observe from the table that a lower average agreement rate leads to better performance. By contrast, $N=6$ yields a higher consistency ratio but achieves the worst results.

We therefore adopt $N=8$ as the default rollout setting in all main experiments. Due to GPU memory constraints, we did not explore $N=16$ or more and leave the study of larger rollout sizes and their potential interaction with the $N$-vote inference scale to future work.

\section{GLPO Algorithm}\label{algorithm}

Algorithm~\ref{alg:glpo} summarizes the GLPO training procedure. GLPO extends GRPO by incorporating a group-level voting reward into the advantage computation. 
For each prompt, the policy samples $K$ responses and computes their individual rewards. 
A group-level response is then obtained through frequency-based voting, and its reward is used as a reference signal. 

\end{document}